\documentclass{article}

\usepackage{arxiv}

\usepackage[utf8]{inputenc} % allow utf-8 input
\usepackage[T1]{fontenc}    % use 8-bit T1 fonts
\usepackage{hyperref}       % hyperlinks
\usepackage{url}            % simple URL typesetting
\usepackage{booktabs}       % professional-quality tables
\usepackage{amsfonts}       % blackboard math symbols
\usepackage{graphicx}
\usepackage{doi}

\usepackage{multirow}
\usepackage{amsmath,amssymb,amsfonts}
\usepackage{amsthm}
\usepackage{mathrsfs}
\usepackage[title]{appendix}
\usepackage{graphicx}
\usepackage{epstopdf}
\usepackage{listings}
\usepackage[numbers,sort&compress]{natbib}
\usepackage{tablefootnote}
\usepackage{textcomp}%
\usepackage{manyfoot}%
\usepackage{subcaption}

\usepackage{lmodern}
\usepackage{booktabs}
\usepackage{hyperref}
\usepackage{xcolor}

\newcommand{\seqToSeq}{sequence-to-sequence}
\newcommand{\GRU}{\text{GRU}}
\newcommand{\RNN}{RNN}
\newcommand{\transformer}{\text{Transformer}}
\newcommand{\DATA}{DATA}
\newcommand{\QUERY}{QUERY}
\newcommand{\INDEX}{INDEX}
\newcommand{\KEY}{KEY}
\newcommand{\NOTRELEVANT}{NOT\_RELEVANT}
\newcommand{\Emb}{\text{Embed}}
\newcommand{\Lin}{\text{Lin}}
\newcommand{\Softmax}{\text{Softmax}}
\newcommand{\LogSoftmax}{\text{MLogSoftmax}}
\newcommand{\argmax}{\text{argmax}}
\newcommand{\selectiveRead}{selective read}
\newcommand{\permnetd}{\textit{PermNetD}}
\newcommand{\permneti}{\textit{PermNetI}}

\newcommand{\num}[1]{#1}
\newcommand{\lwr}[1]{${}_{\text{#1}}$}
\newcommand{\dsDTen}{PD\lwr{10}}
\newcommand{\dsITen}{PI\lwr{10}}
\newcommand{\dsDTwenty}{PD\lwr{20}}
\newcommand{\dsDForty}{PD\lwr{40}}
\newcommand{\dsDHundred}{PD\lwr{100}}
\newcommand{\dsITwenty}{PI\lwr{20}}
\newcommand{\dsDOneToTen}{PD\lwr{1\nobreakdash-10}}
\newcommand{\dsIOneToTen}{PI\lwr{1\nobreakdash-10}}
\newcommand{\dsIForty}{PI\lwr{40}}
\newcommand{\dsIHundred}{PI\lwr{100}}
\newcommand{\DICT}{PI\lwr{DICT}}

\newcommand{\exampleStr}[1]{\textit{"#1"}}

\newcommand{\parenthesis}[1]{\left(#1\right)}

\definecolor{codegreen}{rgb}{0,0.6,0}
\definecolor{codegray}{rgb}{0.5,0.5,0.5}
\definecolor{codepurple}{rgb}{0.58,0,0.82}
\definecolor{backcolour}{rgb}{0.95,0.95,0.92}

\lstdefinestyle{mystyle}{
	backgroundcolor=\color{backcolour},
	commentstyle=\color{codegreen},
	keywordstyle=\color{magenta},
	numberstyle=\tiny\color{codegray},
	stringstyle=\color{codepurple},
	basicstyle=\ttfamily\footnotesize,
	breakatwhitespace=false,
	breaklines=true,
	captionpos=b,
	keepspaces=true,
	showspaces=false,
	showstringspaces=false,
	showtabs=false,
	tabsize=4,
}

\title{Neural architectures for resolving references in program code}

\author{
	\href{https://orcid.org/0009-0008-9782-3869}{\includegraphics[scale=0.06]{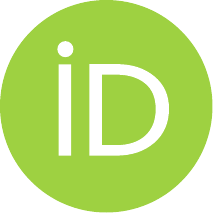}\hspace{1mm}Gergő~Szalay} \\
	Faculty of Informatics\\
	ELTE Eötvös Loránd University\\
	Budapest, H-1117 \\
	\texttt{d5ij3p@inf.elte.hu} \\
	\And
	\href{https://orcid.org/0009-0004-5067-1686}{\includegraphics[scale=0.06]{images/orcid.pdf}\hspace{1mm}Gergely Zsolt Kovács}\\
	Faculty of Informatics\\
	ELTE Eötvös Loránd University\\
	Budapest, H-1117 \\
	\texttt{kovacsg@inf.elte.hu} \\
	\And
	\href{https://orcid.org/0009-0008-2929-7422}{\includegraphics[scale=0.06]{images/orcid.pdf}\hspace{1mm}Sándor Teleki}\\
	Faculty of Informatics\\
	ELTE Eötvös Loránd University\\
	Budapest, H-1117 \\
	\texttt{i3jxte@inf.elte.hu} \\
	\AND
	\href{https://orcid.org/0000-0003-3431-0667}{\includegraphics[scale=0.06]{images/orcid.pdf}\hspace{1mm}Balázs Pintér}\\
	Faculty of Informatics\\
	ELTE Eötvös Loránd University\\
	Budapest, H-1117 \\
	\texttt{pinter@inf.elte.hu} \\
	\And
	\href{https://orcid.org/0000-0002-9503-9623}{\includegraphics[scale=0.06]{images/orcid.pdf}\hspace{1mm}Tibor Gregorics}\\
	Faculty of Informatics\\
	ELTE Eötvös Loránd University\\
	Budapest, H-1117 \\
	\texttt{gt@inf.elte.hu} \\
}

\renewcommand{\headeright}{Preprint}
\renewcommand{\undertitle}{Preprint}
\renewcommand{\shorttitle}{Neural architectures for resolving references in program code}

\hypersetup{
	pdftitle={Neural architectures for resolving references in program code},
	pdfsubject={cs.AI},
	pdfauthor={Gergő Szalay, Balázs Pintér},
	pdfkeywords={reference rewriting, program code, copying mechanism, sequence-to-sequence architecture},
}

\begin{document}

\keywords{reference rewriting, program code, copying mechanism, sequence-to-sequence architecture}

\maketitle

\begin{abstract}
	Resolving and rewriting references is fundamental in programming
	languages. Motivated by a real-world decompilation task, we abstract reference
	rewriting into the problems of direct and indirect indexing by permutation. We
	create synthetic benchmarks for these tasks and show that well-known
	sequence-to-sequence machine learning architectures are struggling on these
	benchmarks. We introduce new sequence-to-sequence architectures for both
	problems. Our measurements show that our architectures outperform the
	baselines in both robustness and scalability: our models can handle examples
	that are ten times longer compared to the best baseline. We measure the impact
	of our architecture in the real\nobreakdash-world task of decompiling switch statements,
	which has an indexing subtask. According to our measurements, the extended
	model decreases the error rate by 42\%. Multiple ablation studies show that
	all components of our architectures are essential.
\end{abstract}

\section{Introduction}
\label{sec:intro}

Resolving references is fundamental in programming languages. A
straightforward example is data flow analysis, where variable definitions, uses,
and modifications are tracked throughout the program: we need to know which
names refer to the same variable. An illustrative subcase is constant
propagation \cite{wegman1991constant}, where, in the simplest case, we find the
variables that refer to constants, and replace them with the constant. Some
other examples include function inlining, macro expansion, or generic
instantiation.

Referring to other parts of the text is also a fundamental part of natural languages,
particularly through linguistic constructs like anaphora. For example, in the
sentence \exampleStr{The trophy doesn’t fit in the brown suitcase because it’s
	too big.} \cite{levesque2012winograd}, \exampleStr{it} could refer to either
the trophy or the suitcase.

The two main NLP tasks involving anaphoras are anaphora resolution
\cite{mitkov2014anaphora} and anaphora rewriting
\cite{quan2019gecor,tseng2021cread}. In anaphora resolution, we would identify
whether \exampleStr{it} stands for \exampleStr{trophy} or \exampleStr{suitcase}.
In anaphora rewriting, we would also replace it in the sentence, so our result
would be \exampleStr{The trophy doesn’t fit in the brown suitcase because the
	trophy is too big.}

Although the two problems appear to be the same,
anaphora rewriting in natural languages and reference rewriting in programming
languages are different in two important ways. In reference rewriting, the
references are less ambiguous, although the problem can still be hard (e.g., the
constant propagation problem is undecidable in general). The second difference
is that in reference rewriting, there are typically many more references and
referents.

\begin{figure}
	\centering
	\includegraphics[width=0.7\textwidth]{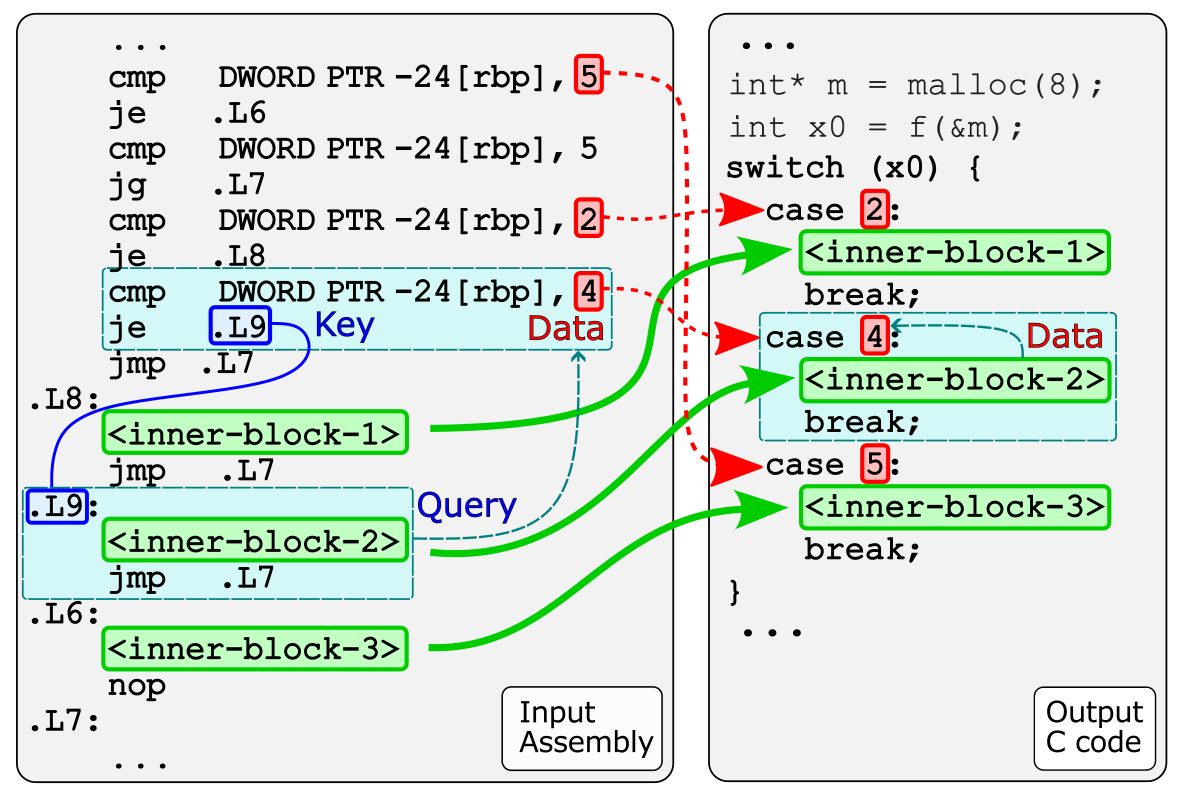}
	\caption{\label{fig:motivation} An example of decompiling a switch statement
		as a reference rewriting task. The assembly snippet on the left is
		decompiled into the C snippet on the right. Although the branches (green)
		appear in the same order on both sides, the order of the case labels (red)
		differs: they need to be reordered during decompilation. The connecting link
		between a branch and its corresponding case label is the jump label in the
		assembly. We can regard the jump label for the branch as the query
		(reference), the jump label for the case label as the key, and the case
		label itself as the data (referent).}
\end{figure}

We encountered a special case of reference rewriting in neural decompilation
when decompiling switch statements from assembly to C
(Figure~\ref{fig:motivation}). In the compiled assembly, the evaluation of the
switch conditions (comparing against the values of the case labels) is separated
from the branches for more efficient computation. The compiler may also
rearrange the order in which these labels are checked in order to further
increase runtime speed. As a result, even though the order of the branches is
the same, the order of the case labels differs between assembly and C. A neural
decompilation tool needs to be able to rewrite the case labels in the correct
order. The order is determined by the jump labels in the assembly: we can regard
the jump label for the branch as the query (reference), the jump label for the
case label as the key, and the case label itself as the data (referent). This
application is evaluated in Section~\ref{sec:realWorldIndirect} and a real-world
example can be found in Appendix~\ref{sec:realWorldExample}.

To study the general problem and focus on its fundamental aspects, we are going
to introduce the abstract problem of \emph{indexing by permutation}.
Furthermore, we distinguish between \emph{direct} and \emph{indirect} indexing
by permutation. In direct indexing by permutation, we have two input sequences:
a \emph{data} sequence and an \emph{index} sequence. The task is to rewrite
(reorder) the data sequence in the order of the index sequence.

An example of this problem would be
\begin{equation}
	a\ c\ b,\ 3\ 1\ 2 \to b\ a\ c,
\end{equation}
where $a$, $c$, and $b$ are the data, and $3$, $2$, $1$ are the indices. A
limitation of this problem formulation is that the position each index token
represents has to be encoded into the representations of the tokens themselves,
so the set of usable indices cannot change nor expand once the model has been
trained.

Indirect indexing is a generalization of direct indexing and is more similar to
the general reference rewriting problem: instead of indices, it uses keys. We
have the following inputs: a sequence of \emph{data} - \emph{key} pairs, and a
\emph{query} sequence, which is a permuted sequence of the keys. In this case,
since the key tokens themselves do not hold the position information, both the
data and key tokens can be freely changed and expanded even after training.

An interesting non-programming example of indirect indexing is substituting the
citations in a paper with the author names and titles. Here, the data would be
the bibliographic entries, the keys would be the citation keys, and the query
sequence would be the citation keys in the body of the paper.

Two examples of this problem are
\begin{equation}
	\begin{aligned}
		& a\ 1\ b\ 2\ c\ 3,\ 1\ 3\ 2 \to a\ c\ b \\
		& a\ 2\ b\ 1\ c\ 3,\ 1\ 3\ 2 \to b\ c\ a, \\
	\end{aligned}
\end{equation}
where in the first part $a\ 1\ b\ 2\ c\ 3$ contains the data-key pairs (e.g. $a\ 1$), and the second part $1\ 3\ 2$ is the query sequence.

In this paper, we address the direct and indirect permutation problems. We
compile a benchmark dataset and evaluate existing architectures. We find that
they are incapable of solving these problems as the number of data items
increases even to only 10-20 items. After reasoning about the problem and introducing
new deep learning architectures tailored to its stucture, we find that these new
architectures can reliably solve it, even for a larger amount (40-100) of items.

Our contributions are as follows:
\begin{itemize}
	\item We introduce direct and indirect indexing by permutation as abstracted reference rewriting subtasks in sequence-to-sequence tasks for program code.
	\item We introduce neural architectures to address these two problems.
	\item We show that our architectures can solve these problems for significantly larger input sizes compared to the baselines.
\end{itemize}

\section{Related work}
\label{sec:relatedWorks}

Referencing parts of the input sequence has been explored before. Perhaps the
most well-known example is the pointer network \cite{pointer}, which can point
to elements of the input sequence to produce an output sequence which consists of
elements of the input. This way the architecture can have a dynamic vocabulary:
it can solve problems where the output sequence consists of elements of the
input sequence arranged in a specific order (with possible duplications), and
these are not known beforehand. For example, when finding a planar convex hull,
the input points change across different problems, so a fixed output vocabulary
would not work.

\citet{copy} integrated the pointing mechanism into the sequence-to-sequence
architecture to give it the capability of \emph{copying} tokens from the input.
Their CopyNet network can operate in generate mode and in copy mode: each output
token is either generated from the fixed vocabulary like in a vanilla
sequence-to-sequence network, or it is copied from the input. The logits of the
next token are computed separately in the two modes and added together, thus
obtaining the final probability of the token in the distribution.

CopyNet can handle rare and out-of-vocabulary tokens efficiently and can provide
precise copying of sequences from the input to the output. However, the network
is designed to preserve the sequence order during copying and, according to our
measurements, it has difficulties when solving indexing by permutation problems.

\citet{see2017get} introduce the pointer-generator network for text
summarization, which is very similar to CopyNet, the main difference being that
they calculate an explicit switch probability $p_{gen}$ to decide between the
generate mode and copy mode.

An interesting application of the pointer-generator network to references in
natural language is the GECOR system \cite{quan2019gecor}, which rewrites
anaphora to their antecedents in dialogue queries by copying them from their
wider context.

\section{Methods}
\label{sec:Methods}

In this section we introduce two different neural network architectures
created for the two (direct and indirect) indexing by permutation problems. Both
are based on a \seqToSeq{} architecture with a bidirectional \GRU{} encoder \cite{gru, bidirectional} and
a one-directional \GRU{} decoder.

From here on, we will refer to tokens representing data with \DATA{}, indices with \INDEX{}, keys with \KEY{}, permuted keys with \QUERY{}, and tokens unrelated to the problem with \NOTRELEVANT{}.

\subsection{Architecture for the direct permutation problem}
\label{sec:directModel}

\begin{figure}[!h]
	\centering
	\includegraphics[width=\textwidth]{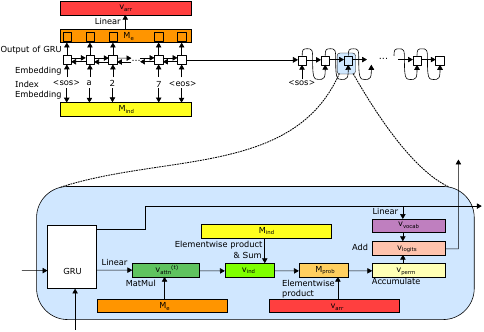}
	\caption{
		\label{fig:directModel}The architecture created for the direct permutation problem.
		The encoder and the decoder can be seen in the top left and the top right corner, respectively.
		The inner structure of the decoder is shown at  the bottom.
		The input is processed by two different embedding layers.
		One of these maps the tokens to a potential indexing distribution, the other one provides an input to the bidirectional encoder \GRU{}.
		The output of the encoder \GRU{} is given to a linear layer to specify the beginning of the \DATA{} sequence.
		The hidden representation created by the encoder \GRU{} is used by the decoder.
		The new hidden vector given by the \GRU{} of the decoder is used to generate the logits for a fixed vocabulary, through a linear layer.
		The new hidden vector is also combined with the encoder outputs to select the current indexing position.
		This distribution is then element-wise multiplied with the index embedding values and accumulated to produce a single index distribution.
		This is then element-wise multiplied further with the sequence-beginning distribution, resulting in the logits that select the appropriate locations from the input.
		Both logits are summed to get the final output of the decoder step.
	}
\end{figure}

For the direct permutation problem, the maximal length of the \DATA{} sequences is known since the indices cannot be modified after training.
Another requirement is that the \DATA{} must be in a contiguous sequence within the input, otherwise the \DATA{} and \INDEX{} tokens may not be distinguishable.
We created our model (\permnetd{}) for direct permutation based on these priors.
An overview of the architecture can be seen in Figure~\ref{fig:directModel}.

The main idea behind this architecture is inspired by array indexing in \textit{C}-style programming languages.
In the C language, we need a pointer to the first element of the array, and then the index defines the offset of the requested element.
To be more specific, the offset is the index multiplied by the size of an element, which is $1$ in our case, since we store standalone tokens in the array.
Described below is a \seqToSeq{} architecture extended by this mechanism in a differentiable way.

The input is processed by an embedding layer and a bidirectional \GRU{}, whose last hidden state is passed to the decoder (Eq.~\ref{eq:dirEncoder}).
The encoder outputs are mapped to a single value for each input token by a linear layer (Eq.~\ref{eq:dirArr}).
This results in a vector which, according to our motivation, should point to the first element of the array with a large weight.

\begin{align}
	M_e, \mathbf{v_{last}} &= \GRU_{Enc}(\Emb(\mathbf{v_{inp}})) \label{eq:dirEncoder}\\
	% batch x max_length x 2*hidden ; 2 x batch x hidden
	\mathbf{v_{arr}} &= \Lin(M_e) \label{eq:dirArr}
	% batch x max_length x 1
\end{align}

A second embedding layer is applied to the input, with an embedding size that is equal to the maximal array length.
This embedding can be thought of as a matrix mapping tokens to (one-hot embedding-like) indices.

\begin{equation}
	\label{eq:indexEmbedding}
	M_{ind} = \Emb_{ind}(\mathbf{v_{inp}})
	% batch x max_length x max_perm
\end{equation}

The decoder generates the output tokens one-by-one, processing the previous
token in each step. We applied the same token embedding for both \RNN{}s. After
the previous token is processed (Eq.~\ref{eq:dirDec}), the new hidden state is processed by a linear layer and
multiplied by the outputs of the encoder (Eq.~\ref{eq:dirSelect}). This can be interpreted as pointing to
the next \INDEX{} token in the input. A similar behavior is used by the copying
mechanism~\cite{copy}.
We use these values as weights and multiply them with the index values given by
the second embedding layer ($M_{ind}$). We sum the weighted index values,
creating a single index vector ($\mathbf{v_{ind}}$, Eq.~\ref{eq:dirInd}).

\begin{align}
	\textvisiblespace, \mathbf{v_{hidd}^{(t)}} &= \GRU_{Dec}(\Emb(p_{out}^{(t-1)}), \mathbf{v_{hidd}^{(t-1)}}) \label{eq:dirDec} \\
	% _ ; 1 x batch_size x 2*hidden
	\mathbf{v_{attn}^{(t)}} &= M_e \cdot \Lin(\mathbf{v_{hidd}^{(t)}}) \label{eq:dirSelect}\\
	% batch x max_length x 1
	\mathbf{v_{ind}^{(t)}}^T &= \sum_\text{column-wise} (M_{ind} \odot \mathbf{v_{attn}^{(t)}}) \label{eq:dirInd}
	% bacth x 1 x max_perm
\end{align}

We multiply the created index vector with the vector "pointing to the first element of the array" (Eq.~\ref{eq:dirProb}).
The resulting matrix contains weights in position $(i, j)$ representing the logit of the element at index $j$ for an array starting at position $i$.
Since there are multiple logits in this matrix pointing to the same input token, we accumulate these values position-wise.
As a result, we get a weight for all the input positions, which should be summed up for the positions, where the same token is presented (Eq.~\ref{eq:dirLogit}).
This second accumulation process is noted with $\tilde\sum$.

\begin{align}
	M_{prob}^{(t)} &= \mathbf{v_{arr}} \odot \mathbf{v_{ind}^{(t)}}^T \label{eq:dirProb} \\
	% batch x max_length x max_perm
	\mathbf{v_{perm}^{(t)}} &= \tilde\sum\left(\sum_{\text{antidiagonal-wise}} M_{prob}^{(t)}\right) \label{eq:dirLogit}
	% batch x vocab
\end{align}

We also try to predict the tokens from a given vocabulary, applying a linear layer to the hidden state of the current step.

\begin{equation}
	\mathbf{v_{vocab}^{(t)}} = \Lin(\mathbf{v_{hidd}^{(t)}}) \label{eq:vocab}
	% 1 x batch x vocab
\end{equation}

The result of the model is the sum of these two logits.

\begin{align}
	\mathbf{v_{logits}^{(t)}} = \mathbf{v_{perm}^{(t)}} + \mathbf{v_{vocab}^{(t)}}\\
	% batch x vocab
	p_{out}^{(t)} = \argmax(\mathbf{v_{logits}^{(t)}})
\end{align}

During the training we applied weight decay for $\Emb_{ind}$ (used in Eq.~\ref{eq:indexEmbedding}), so the index embedding layer could eliminate the weights of the non-index tokens.

\subsection{Architecture for the indirect permutation problem}
\label{sec:indirectModel}

\begin{figure}[!h]
	\centering
	\includegraphics[width=\textwidth]{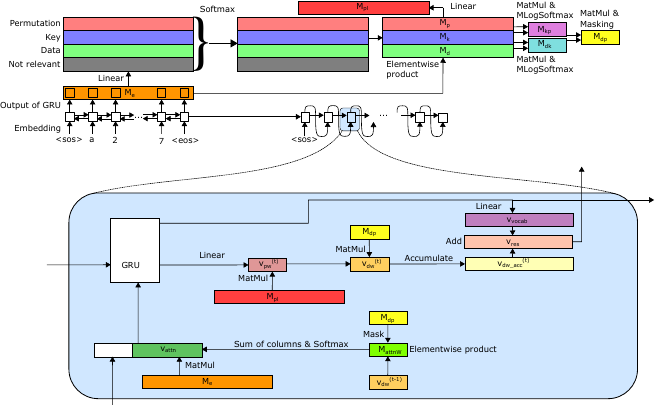}
	\caption{
		\label{fig:indirectModel}The architecture created for the indirect permutation problem with the attention mechanism (detailed in Section~\ref{sec:attention}).
		The encoder can be seen on the top and the left of the figure, while the structure of the decoder can be seen at the bottom.
		The encoder processes the input token-wise with an embedding and a bidirectional \GRU{} layer.
		The outputs of this \GRU{} are given to a linear layer to produce a distribution that determines whether the given token is a \DATA{}, a \KEY{}, a \QUERY{}, or \NOTRELEVANT{}.
		Using these values, associations are created from the \QUERY{} tokens to the \KEY{} tokens and from the \KEY{} tokens to the \DATA{} tokens respectively.
		These connections are then matrix-multiplied to produce the association from \QUERY{} to \DATA{}.
		The hidden vector created by the \GRU{} of the encoder is given to the decoder for further processing.
		In each step, the decoder creates a new hidden vector, which is forwarded to a linear layer to predict the logits from a fixed vocabulary.
		Simultaneously, the new hidden vector is multiplied by the \QUERY{} weights created by the encoder and further multiplied by the association matrix created in the encoder to predict the output.
		The two outputs are then summed, resulting in the output of the decoder step.
	}
\end{figure}

In the case of the indirect problem, there is no need to learn some specific
meaning for a set of tokens, thus there is no limitation to the length of the \DATA{} sequence.
Instead, the model (\permneti{}) needs to identify the \DATA{}-\KEY{} pairs, the \QUERY{} sequence,
and the permutation transforming the \QUERY{} tokens to the \KEY{} tokens,
and apply this permutation to the \DATA{} tokens.
Another way to think about this problem is
that the model needs to use the given \DATA{}-\KEY{} pairs to map
the \QUERY{} tokens to the specific data tokens. This is the main idea behind
the created architecture. Figure~\ref{fig:indirectModel} shows the layout of the model, including
the attention mechanism, which will be discussed in Section~\ref{sec:attention}.

The architecture is based on a \seqToSeq{} \RNN{} model. The encoder is a
bidirectional \GRU{}, which processes the embedded input. We save the encoder
outputs as $M_e$ and its last hidden state as $\mathbf{v_{last}}$.

\begin{equation}
	M_e, \mathbf{v_{last}} = \GRU_{Enc}(\Emb(\mathbf{v_{inp}}))
	% batch x max_length x 2*hidden ; 2 x batch x hidden
\end{equation}

The encoder outputs are processed by a linear layer and a softmax function,
mapping each output to a four-valued probability distribution ($M_{label}$, Eq.~\ref{eq:indLabel}).
This specifies four probabilities for each token: whether they belong to the
\DATA{}, \KEY{}, \QUERY{} or \NOTRELEVANT{} categories. Based on this, we
separate the values from each of the \DATA{}, \KEY{}, and \QUERY{} categories
into three vectors and multiply them element-wise with the outputs of the
encoder. This results in three matrices: $M_{d}$, $M_{k}$, and $M_p$.

\begin{align}
	M_{label} &= \Softmax(\Lin(M_e)) \label{eq:indLabel} \\
	% batch x max_length x 4
	\label{eq:Md}M_d &= M_e \odot M_{label}[1]\\
	% batch x max_length x 2*hidden
	\label{eq:Mk}M_k &= M_e \odot M_{label}[2]\\
	\label{eq:Mp}M_p &= M_e \odot M_{label}[3]
\end{align}

Using these weighted encoder output values, we perform two matrix multiplications followed by a modified log-softmax (Eq.~\ref{eq:indDK} and Eq.~\ref{eq:indKP}).
The latter is necessary to obtain a probability-like distribution of the values. Compared to a basic softmax function, this approach uses a logarithm to achieve better gradient flow, which is important in the middle of the model.
Because the original log-softmax function produces negative values, the following matrix multiplication would result in a monotonically decreasing function.
To prevent this, we transform the result of the softmax function (the argument of the logarithm) to be between $1$ and $e$, thus preserving the monotonically increasing behavior of the transformation (Eq.~\ref{eq:indLogSoftmax}).
It is easy to see that in the case of one value being significantly larger than the others, the result of this transformation is identical to the basic softmax function.
The resulting two square matrices ($M_{dk}$, $M_{kp}$) have dimensions equal to the input length.
They represent the connection between the \KEY{} and the \DATA{} tokens, and between the \QUERY{} and the \KEY{} tokens respectively.
Knowing these relations, we can get the connections between the \QUERY{} and \DATA{} tokens through the \KEY{} tokens by multiplying the two matrices, resulting in $M_{dp}$ (Eq.~\ref{eq:indDP}).
To prevent cases when a token in a given position has the largest correlation with itself, we set the main diagonals of these matrices to $0$ (we also do this to the values corresponding to padding tokens).

\begin{align}
	\LogSoftmax(x) &:= \ln(1 + (e - 1) \cdot \Softmax(x)) \label{eq:indLogSoftmax} \\
	M_{dk} &= \LogSoftmax(M_d \cdot M_k^T) \label{eq:indDK} \\
	% batch x max_length x max_length
	M_{kp} &= \LogSoftmax(M_k \cdot M_p^T) \label{eq:indKP} \\
	M_{dp} &= M_{dk} \cdot M_{kp} \label{eq:indDP}
	% batch x max_length x max_length
\end{align}

The decoder generates the output token by token, processing the previous token in each step.
The one-directional \GRU{} calculates the new hidden state ($\mathbf{v_{hidd}^{(t)}}$) based on the token created at the previous step ($p_{out}^{(t-1)}$, Eq.~\ref{eq:indDec}).
The logits for the tokens from the vocabulary is predicted from the hidden state ($\mathbf{v_{vocab}^{(t)}}$).

\begin{align}
	\textvisiblespace, \mathbf{v_{hidd}^{(t)}} &= \GRU_{Dec}(\Emb(p_{out}^{(t-1)}), \mathbf{v_{hidd}^{(t-1)}}) \label{eq:indDec}\\
	% _ ; 1 x batch x 2*hidden
	\mathbf{v_{vocab}^{(t)}} &= \Lin(\mathbf{v_{hidd}^{(t)}})
	% batch x vocab
\end{align}

Apart from this, we use linear layers on the new hidden state and the outputs of the encoder weighted by the \QUERY{} ($M_p$), and multiply the resulting vector and matrix (Eq.~\ref{eq:indPW}).
This results in a vector weighting the \QUERY{} tokens for the next prediction.
We now multiply this vector with the $M_{dp}$ matrix to get a weighting on the \DATA{} tokens ($\mathbf{v_{dw}^{(t)}}$).

\begin{align}
	\mathbf{v_{pw}^{(t)}} &= \Lin(M_p) \cdot \Lin(\mathbf{v_{hidd}^{(t)}}) \label{eq:indPW} \\
	% batch x max_length x 1
	\mathbf{v_{dw}^{(t)}} &= M_{dp} \cdot \mathbf{v_{pw}^{(t)}}
	% batch x max_length x 1
\end{align}

At this point we know which tokens are relevant from the input, however, a token could have multiple occurrences in the input.
Since the specific positions of tokens do not matter for predicting the output, we accumulate the weights of positions having the same token.
This accumulated vector is then added to the prediction of the vector $\mathbf{v_{vocab}^{(t)}}$.
The predicted token is the one with the highest value in this vector.

\subsubsection{Attention mechanism for the indirect permutation problem}
\label{sec:attention}

Motivated by the attention of the copying mechanism (\selectiveRead{}, \cite{copy}) we designed an attention mechanism to increase the effectiveness of the architecture described in Section~\ref{sec:indirectModel}.
The \selectiveRead{} of the copying mechanism uses the logits from the previous step as the weights for the attention.
However, in our case the attention should be focused on the \QUERY{} tokens instead of the \DATA{} tokens, thus we need to transform the logits back to the \QUERY{} tokens.

At the beginning of the decoder step, the predicted token is known from the previous step.
If the predicted token is part of the input sequence, we can select the data weights ($\mathbf{v_{dw}^{(t-1)}}$) related  to the position of the token.
The remaining weights are set to zero ($\mathbf{v_\text{selected}^{(t-1)}}$).

Since the data weights ($\mathbf{v_{dw}^{(t-1)}}$) are calculated from the \QUERY{} weights ($\mathbf{v_{pw}^{(t-1)}}$) by multiplying them with the $M_{dp}$ matrix, one could try to get the inverse relation by applying a matrix multiplication with the inverse of $M_{dp}$.
Unfortunately, this is not possible, since $M_{dp}$ is singular in most cases and, even if it was regular (invertible), the calculation of the inverse matrix would be expensive and numerically unstable.
There are also similar problems with the pseudo-inverse calculation.

Let us note that only some weights should be mapped back to the \QUERY{} weights.
Because the rows of $M_{dp}$ represent mapping a weighting from \QUERY{} to \DATA{}, each row is related to one output logit from the previous step.
Knowing the selected tokens ($\mathbf{v_\text{selected}^{(t-1)}}$) we can weight these rows with the corresponding logits (Eq.~\ref{eq:indAttnW}).
Since each value in a row shows how relevant a given \QUERY{} token was in the previous step, this weighting shows for each element of each row how significant it was in the last step.
After getting these weighted rows, we can sum them up to get the accumulated attention weights for each token (Eq.~\ref{eq:indAttn}).
These weights can now be used in the same way as in the original attention mechanism of~\citet{bahdanau}.

\begin{align}
	M_{attnW}^{(t)} &= M_{dp} \odot \mathbf{v_\text{selected}^{(t-1)}} \label{eq:indAttnW}\\
	% batch x max_length x max_length
	\mathbf{v_{attn}^{(t)}} &= M_e^T \cdot \text{Softmax}\parenthesis{\sum_{\text{column-wise}} M_{attnW}^{(t)}}^T \label{eq:indAttn}
	% batch x 2*hidden x 1
\end{align}

\subsubsection{On the compatibility of the attention mechanisms}
\label{sec:compatibility}
The attention mechanism described in Section~\ref{sec:attention} calculates the attention weights based on the previous output logits.
Hence, the "focus" of this attention does not interfere with the attention mechanism of~\citet{bahdanau}, which calculates its weight from the current hidden state.
In the case of the \selectiveRead{} of the copying mechanism, however, it is not that simple, since the \selectiveRead{} also uses the previous logits as weights.

On one hand, extending the logit prediction of the copying mechanism with our architecture is  really straightforward, since both methods apply a weighting to the input tokens, so the result of these two methods could be merged as well.
On the other hand, using the logits of such a merged result can cause ambiguity.
Both attention mechanisms calculate the attention weights from the previous logits, but they use these logits in different ways.
For the \selectiveRead{}, the logits are connected directly to the associated tokens, meanwhile our architecture applies an indirection on the logits to select the connected tokens.
For this reason, we have to eliminate the "race condition" between the two attention mechanisms.
To do this, for each attention, we mask out the logits, where the prediction of the other mechanism had a greater value.
As a result, only one of the attention mechanisms can use each of the output logits, preventing malfunctioning.

\section{Results}
\label{sec:results}

In this section, we first discuss our datasets and metrics for the two
architectures defined in Section~\ref{sec:Methods}. Then we present our results
on the different datasets and compare them to other well-known architectures. We
refer to our architecture for the direct permutation problem as \permnetd{} and
to the one for the indirect permutation problem as \permneti{}. We conclude by
measuring the impact of integrating \permneti{} into a previously developed
architecture as a real-world showcase of its applicability.

\subsection{Datasets}
\label{sec:datasets}

We generated random examples to train, evaluate, and test both the direct and
indirect permutation problems. In the case of the direct permutation problem,
every example starts with \DATA{} tokens, then has a \texttt{<sep>} token and
ends with \INDEX{} tokens. The examples generated for the indirect permutation
problem start with \DATA{}-\KEY{} token pairs and, after a separating
\texttt{<sep>} token, end with \QUERY{} tokens. We enclose all input sequences
within an \texttt{<sos>}, \texttt{<eos>} token pair.

We created datasets for both problems, each with a different number of \DATA{}
tokens, which we will refer to as the length of an example: \dsDTen{},
\dsDTwenty{}, \dsDOneToTen{} (from 1 to 10) for the direct permutation problem,
and \dsITen{}, \dsITwenty{}, \dsIOneToTen{} for the indirect one. The
\dsDOneToTen{} and \dsIOneToTen{} benchmarks contain examples of different
lengths (from 1 to 10) to test the generalization capabilities of the models.

\begin{lstlisting}[caption={An example from the \dsDTen{} dataset.}]
	Input:
	<sos> d i j h e g f b c a <sep> 3 2 7 6 5 1 4 8 0 9 <eos>
	Output:
	<sos> h j b f g i e c d a <eos>
\end{lstlisting}

\begin{lstlisting}[caption={An example from the \dsITen{} dataset.}]
	Input:
	<sos> d 8 b 9 g 6 j 10 e 5 i 4 a 3 c 2 f 1 h 7 <sep> 2 10 4 3 1 8 5 9 6 7 <eos>
	Output:
	<sos> c j i a f d e b g h <eos>
\end{lstlisting}

All of our datasets contain 300\,000 examples equally distributed for training, evaluation, and testing.

For further evaluation, we created two additional benchmarks for the direct, and
three additional benchmarks for the indirect permutation problem. Four of them
are similar to the original datasets, but with lengths of 40 (\dsDForty{},
\dsIForty{}) and 100 (\dsDHundred{}, \dsIHundred{}). The fifth additional
dataset is based on \dsITen{}, but its \QUERY{} tokens are not necessarily
unique. This fifth dataset will be referred to as \DICT{}, since it acts like
indexing into a dictionary multiple times.

\begin{lstlisting}[caption={An example from the \DICT{} dataset.}]
	Input:
	<sos> j 10 b 7 c 3 i 2 d 1 a 8 f 9 g 5 e 6 h 4 <sep> 7 3 9 5 8 6 9 1 2 6 <eos>
	Output:
	<sos> b c f g a e f d i e <eos>
\end{lstlisting}

\subsection{Metrics and evaluation}
\label{sec:metrics}

We used two metrics to measure the performance of the models:
\begin{itemize}
	\item \textbf{Token Accuracy (TA):} The proportion of correctly predicted tokens to the total number of tokens, averaged across all examples.
	\item \textbf{Whole Example Accuracy (WEA):} The proportion of correctly handled examples out of all examples.
\end{itemize}

For evaluation, we took the best result of the baseline architectures and the result of the last epoch for our architectures.
The baseline architectures are \GRU{}~\cite{gru}, \GRU{} with attention~\cite{bahdanau}, copying mechanism~\cite{copy}, and \transformer{}~\cite{attention}.
The hyperparameters of the different models can be found in Appendix \ref{hyperparams}.

\subsection{Results on the direct permutation problem}
\label{sec:resultsDirect}

We compared the baseline architectures with ours (\permnetd{}) on the direct permutation problem.
The results are shown in Table~\ref{tab:directRes}.

To show the scalability of our model, we trained and evaluated it on the \dsDForty{} and \dsDHundred{} datasets.
The results are presented in Table~\ref{tab:directExtra}.

\begin{table}[!ht]
	\centering
	\begin{tabular}{c c c c c c c}
		\toprule
		& \multicolumn{2}{c}{1-10} & \multicolumn{2}{c}{10} &  \multicolumn{2}{c}{20}\\
		& TA & WEA & TA & WEA & TA & WEA\\
		\midrule
		\GRU{} & 21.75\% & 0\% & 17.17\% & 0\% & -- & -- \\
		\GRU{}+\textit{Attn.} & \textbf{99.99\%} & 99.98\% & 24.92\% & 0\% & -- & -- \\
		\textit{CopyNet} & 99.97\% & 99.86\% & \textbf{100}\% & \textbf{100}\% & 94.01\% & 0\%~\protect\footnotemark \\
		\transformer{} & 41.87\% & 0\% & 39.24\% & 0\% & -- & -- \\
		\permnetd{} & \textbf{99.99\%} & \textbf{99.99\%} & \textbf{100\%} & \textbf{100\%} & \textbf{99.95\%} & \textbf{99.62\%} \\
		\bottomrule
	\end{tabular}
	\caption{
		\label{tab:directRes}The results of different architectures on the direct permutation problem.
		If a model could not handle the \dsDTen{} dataset, no further evaluation was made on the \dsDTwenty{} dataset.
		\texttt{TA} marks the token accuracy and \texttt{WEA} marks the whole example accuracy.
	}
\end{table}

\begin{table}[!h]
	\centering
	\begin{tabular*}{0.8\textwidth}{@{\extracolsep{\fill}} c c c c c}
		\toprule
		& \multicolumn{2}{c}{40} & \multicolumn{2}{c}{100}\\
		& TA & WEA & TA & WEA\\
		\midrule
		\permnetd{} & \textbf{99.99\%} & \textbf{99.98\%} & \textbf{99.99\%} & \textbf{99.89\%} \\
		\bottomrule
	\end{tabular*}
	
	\caption{
		\label{tab:directExtra}The results of \permnetd{} on the direct permutation problem.
		\texttt{TA} marks the token accuracy and \texttt{WEA} marks the whole example accuracy.
	}
\end{table}

\subsection{Results on the indirect permutation problem}
\label{sec:resultsIndirect}

Similarly to the direct permutation problem, we compare our architecture (\permneti{}) to the
baselines on the indirect permutation problem. The results are shown in
Table~\ref{tab:indirectRes}.

\begin{table}[!ht]
	\centering
	
	\begin{tabular}{c c c c c c c}
		\toprule
		& \multicolumn{2}{c}{1-10} & \multicolumn{2}{c}{10} &  \multicolumn{2}{c}{20}\\
		& TA & WEA & TA & WEA & TA & WEA\\
		\midrule
		\GRU{} & 19.20\% & 0\% & 16.83\% & 0\% & -- & -- \\
		\GRU{}+\textit{Attn.} & 56.83\% & 17.98\% & 16.35\% & 0\% & -- & -- \\
		\textit{CopyNet} & 99.90\% & 99.32\% & 99.99\% & 99.97\% & 15.93\% & 0\%\\
		\transformer{} & 99.97\% & 19.77\% & 39.15\% & 0\% & -- & -- \\
		\permneti{} & \textbf{99.99\%} & \textbf{99.99\%} & \textbf{100\%} & \textbf{100\%} & \textbf{99.99\%} & \textbf{99.99\%} \\
		\bottomrule
	\end{tabular}
	\caption{
		\label{tab:indirectRes}The results of different architectures on the indirect permutation problem.
		If a model could not handle the \dsITen{} dataset, no further evaluation was made on the \dsITwenty{} dataset.
		\texttt{TA} marks the token accuracy and \texttt{WEA} marks the whole example accuracy.
	}
\end{table}

To show the scalability and generality of our architecture, we trained and tested a model on the \dsIForty{}, \dsIHundred{}, and \DICT{} datasets.
The results of these models are shown in Table~\ref{tab:indirectExtra} and \ref{tab:indirectDict}.

\phantom{}
\footnotetext{
	The learning of the Copying mechanism~\cite{copy} was unstable.
	The highest token accuracies for different training runs range from 25\% to 94\%, but none could solve the task.
	See more details in Section~\ref{sec:discussionDirect}.
}

\begin{table}[!h]
	\centering
	\begin{tabular*}{0.8\textwidth}{@{\extracolsep{\fill}} c c c c c}
		\toprule
		& \multicolumn{2}{c}{40} & \multicolumn{2}{c}{100}\\
		& TA & WEA & TA & WEA\\
		\midrule
		\permneti{} & \textbf{100\%} & \textbf{100\%} & \textbf{99.96\%} & \textbf{98.26\%} \\
		\bottomrule
	\end{tabular*}
	
	\caption{
		\label{tab:indirectExtra}The results of \permneti{} on the indirect permutation problem.
		\texttt{TA} marks the token accuracy and \texttt{WEA} marks the whole example accuracy.
	}
\end{table}

\begin{table}[!h]
	\centering
	\begin{tabular*}{0.8\textwidth}{@{\extracolsep{\fill}} c c c c c}
		\toprule
		& &  \multicolumn{2}{c}{Dict} &\\
		& & TA & WEA \\
		\midrule
		\textit{CopyNet} & & 83.33\% & 0.0\% \\
		\permneti{} & & \textbf{99.94\%} & \textbf{99.72\%} \\
		\bottomrule
		
	\end{tabular*}
	
	\caption{
		\label{tab:indirectDict}The results of copying mechanism~\cite{copy} and the \permneti{} on the \DICT{} permutation problem (defined in Section~\ref{sec:datasets}).
		\texttt{TA} marks the token accuracy and \texttt{WEA} marks the whole example accuracy.
	}
\end{table}

\subsection{Results on a real-world problem}
\label{sec:realWorldIndirect}

We evaluated our architecture for the indirect permutation problem on the
real-world task of decompiling \texttt{switch} statements from assembly to C described in
Section~\ref{sec:intro}. We adopt the architecture introduced
by~\citet{copyDecomp} as our a baseline and extend it with \permneti{}, described
in Section~\ref{sec:indirectModel}.

After training both architectures on examples containing only \texttt{switch}
statements, \permneti{} achieved 91.80\% program decompilation rate (whole
example accuracy), a more than 6 percentage point improvement over the results
of the baseline, which achieved 85.74\%.
The proportion of incorrect cases was reduced by 42\%, showing that the
\permneti{} architecture can be successfully applied to real-world problems.

\section{Discussion}
\label{sec:discussion}

\subsection{Direct permutation problem}
\label{sec:discussionDirect}

The results of the different architectures on the direct permutation problem can
be seen in Table~\ref{tab:directRes}. The \GRU{} with attention
mechanism~\cite{bahdanau} performed well on \dsDOneToTen{}, but failed to
learn the \dsDTen{} dataset. The copying mechanism could handle \dsDTen{}
similarly to \dsDOneToTen{}, however, its learning stability was worse on
\dsDTen{}, and even worse on \dsDTwenty{}, leading to poor results. We performed
several training runs using CopyNet with different hyperparameters, but none of
these could solve the \dsDTwenty{} benchmark. Most of these training runs failed
to achieve a meaningful token accuracy, and the ones which could reach higher
accuracy also struggle with stability, which caused the model to get stuck in the
end, even with the use of gradient clipping.

To show the scalability of \permnetd{}, we trained and evaluated it on the
\dsDForty{} and \dsDHundred{} datasets. As presented in
Table~\ref{tab:directExtra}, the architecture could solve the problem for both
lengths.

\begin{figure}[!ht]
	\centering
	\includegraphics[height=0.5\textheight]{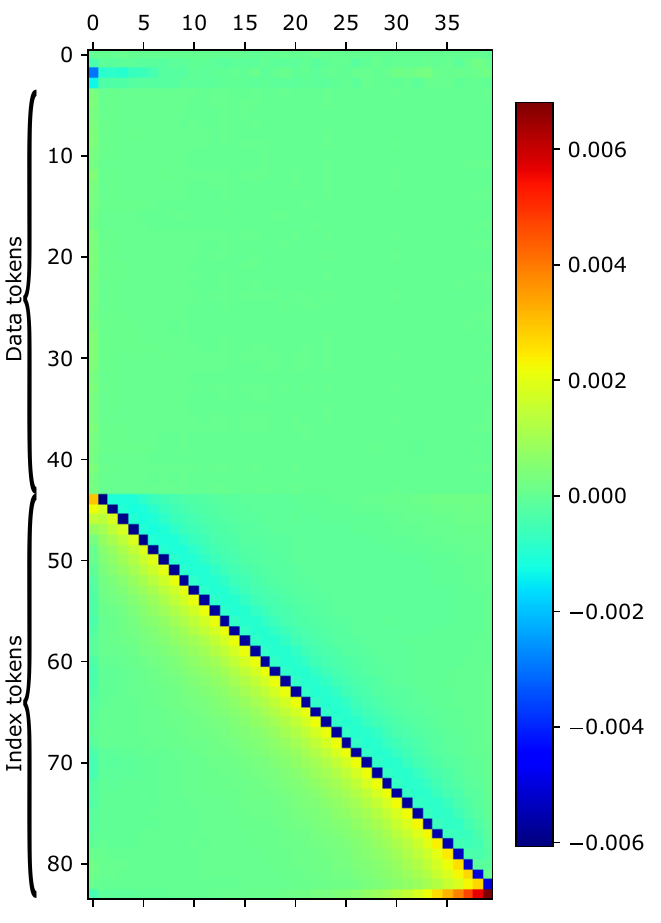}
	\caption{
		\label{fig:emb_weights}The weights of the index embedding of the model
		(Eq.~\ref{eq:indexEmbedding}) for the direct permutation problem after being
		trained on the \dsDForty{} benchmark. The most signicicant values (the
		orange-yellow diagonal) correspond to the correct indexing distribution
		for the relevant tokens. The values for non-index tokens are mostly
		zeroes. }
\end{figure}

In Eq.~\ref{eq:indexEmbedding} an embedding layer is applied to learn
indexing. To observe whether this is really what happened, we visualized the
embedding matrix trained on the \dsDForty{} dataset in Figure~\ref{fig:emb_weights}. Here the \INDEX{}
tokens are located in the last 40 rows in ascending order. We can see that the most
significant (positive) values are in the diagonal of this lower 40-by-40
sub-matrix. This confirms our hypothesis, that the model learns to index with
this mechanism.

\subsection{Indirect permutation problem}
\label{sec:discussionIndirect}

The results in Table~\ref{tab:indirectRes} show that only the copying mechanism~\cite{copy} and \permneti{} can handle examples from the \dsITen{} dataset, but the copying mechanism is inaccurate on longer examples.

Since \permneti{} could solve the basic benchmarks, we were interested in
measuring the scalability of the architecture. To test this, we created examples
with lengths of 40 (\dsIForty{}) and 100 (\dsIHundred{}). The results in
Table~\ref{tab:indirectExtra} show that \permneti{} can also handle these
lengths. This confirms that this architectural addition enables the model to
handle token mappings from \KEY{} to \DATA{} across distance, and the attention
mechanism created for \permneti{} (described in Section~\ref{sec:attention}) can
help to keep the focus of the model.

\begin{figure}[h!]
	\centering
	\includegraphics[height=0.5\textheight]{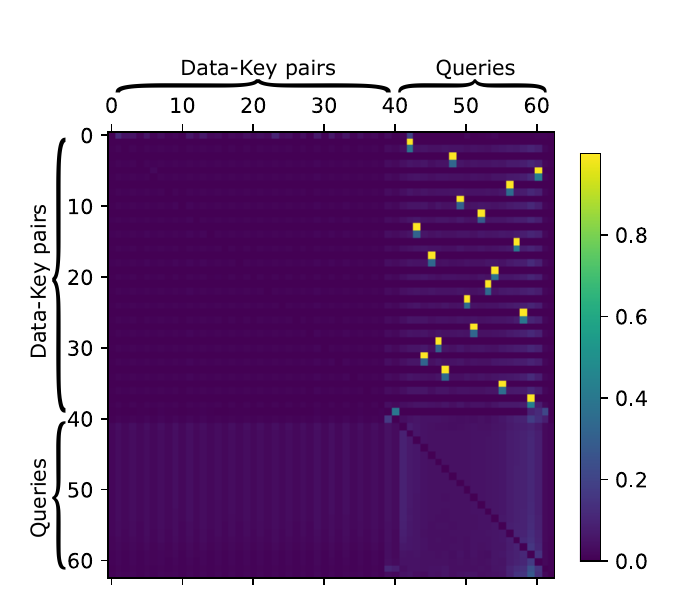}
	\caption{\label{fig:indirectBindingCorrect}
		An example of the $M_{dp}$ matrix (defined in Eq.~\ref{eq:indDP}) calculated by a trained \permneti{} model on a \dsITwenty{} sample.
		This square matrix has dimensions equal to the length of the input sequence.
		The beginning of the sequence contains the \DATA{}-\KEY{} pairs, while the end of the sequence contains the \QUERY{} tokens.
		The matrix connects the \QUERY{} tokens to the \DATA{} tokens, which is visible from the high values in the top right segment.
		These significant values mark the \DATA{} tokens and the related \KEY{} tokens as 2-by-1 areas, with the latter having smaller values.
		This confirms our intuition that the model focuses on the correct \DATA{} tokens.}
\end{figure}

An example of the $M_{dp}$ matrix (defined in Eq.~\ref{eq:indDP}) can be seen in Figure~\ref{fig:indirectBindingCorrect}.
This figure confirms that the model can use the correct reference resolution by this matrix.

To measure the generalizing ability of the architecture, we trained a model on
examples from the \DICT{} dataset. This model reached high accuracy too,
demonstrating that the \permneti{} can handle indexing with non-unique \QUERY{}
values.

\section{Ablation study}
\label{sec:ablationStudy}

The proposed architectures have several components. It is natural to ask which
of them are really necessary to achieve strong performance on the permutation
problems. To answer this question, we investigated the performance of our
architectures after removing components from them.
From now on we will refer to the prediction of the next token from a fixed vocabulary as fixed-vocabulary prediction (Eq.~\ref{eq:vocab}).

For \permnetd{}, we performed the ablation study removing the following
components one by one: the fixed-vocabulary prediction, $M_e$ from
Eq.~\ref{eq:dirSelect} predicting $v_{attn}$, $M_{ind}$, and $v_{arr}$. The
results of these evaluations can be seen in Table~\ref{tab:directWOTrad}.

After removing the fixed-vocabulary prediction, the performance of the
architecture drops significantly as the length of the input increases. When we
removed the multiplication by $M_e$ or $v_{arr}$, both of which try to point to a specific token in
the input, the model could still solve the task. However, for both cases, these
components must be replaced by an extra linear layer whose output size depends
on the input size. This creates a limitation for the use of the model.

In the case of omitting $M_{ind}$, the performance of the model dropped
significantly as the input and output length increased. This shows that learning
the indices is essential for proper functioning.

\begin{table}[h!]
	\centering
	\begin{tabular}{l c c c c c c}
		w/o & \multicolumn{2}{c}{\dsDTen{}} & \multicolumn{2}{c}{\dsDTwenty{}} &  \multicolumn{2}{c}{\dsDForty{}}\\
		& TA & WEA & TA & WEA & TA & WEA\\
		\midrule
		base & 100\% & 100\% & 99.95\% & 99.62\% & 99.99\% & 99.98\% \\
		\midrule
		fixed-vocab. & 100\% & 100\% & 99.92\% & 98.67\% & 86.55\% & 0.17\% \\
		$M_e$ & 99.98\% & 99.93\% & 100\% & 100\% & - & - \\
		$M_{ind}$ & 99.99\% & 99.90\% & 71.51\% & 0.0\% & - & - \\
		$v_{arr}$ & 99.89\% & 99.29\% & 99.99\% & 99.89\% & - & - \\
	\end{tabular}
	\caption{\label{tab:directWOTrad} The results of \permnetd{} after removing parts from it.}
\end{table}

For \permneti{}, we ablated the following components one by one: the
fixed\nobreakdash-vocabulary prediction, the $M_d$\nobreakdash-$M_k$\nobreakdash-$M_p$
matrices, just the $M_d$ matrix, the attention mechanism, and the $M_{pl}$
matrix. The results can be seen in Table~\ref{tab:indirectWO}.

After removing the fixed-vocabulary prediction, the model could not solve the \dsITen{} benchmark, thus this part is essential for correct functionality.

We tried to remove the $M_d$, $M_k$, and $M_p$ matrices all at once and replace them
by a multiplication of the encoder outputs by itself. The performance of the
model dropped drastically.

\begin{figure}[h!]
	\centering
	\includegraphics[height=0.4\textheight]{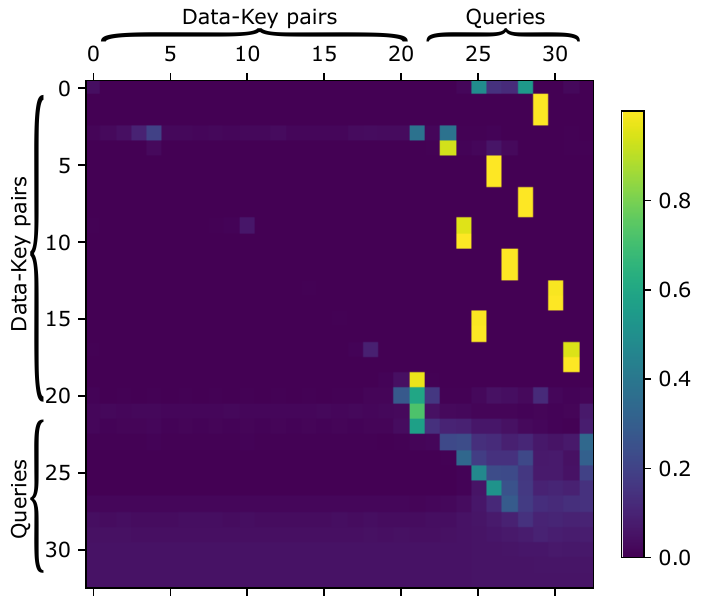}
	\caption{\label{fig:indirectBindingFals}
		An example of the model focusing on the \DATA{}-\KEY{} pairs when removing $M_d$ from \permneti{} on a \dsITen{} example.
		The matrix is similar to Figure~\ref{fig:indirectBindingCorrect}, but there are 10 2-by-1 yellow areas in the picture instead of 10 1-by-1 cells.
		For most of them, the value of the upper part (\DATA{}) is less than or equal to the value of the lower one (\KEY{}).}
\end{figure}

Removing $M_d$ seemingly did not affect the performance, and it also reduced the
number of epochs needed for training. However, upon closer inspection, we
noticed that the functionality of the model changed dramatically. After removing
the $M_d$ matrix, our mechanism was able to focus on the correct \DATA{}-\KEY{}
pair, but was unable to select the data token from this pair (it was focusing on
either the \KEY{}, or both the \KEY{} and the \DATA{}
token).
An example for this behavior can be seen in Figure~\ref{fig:indirectBindingFals}.

To compensate for this, sometimes the fixed-vocabulary prediction learned to
predict low logit values for the \KEY{} tokens and high logits for all
\DATA{} tokens, thus the model predicted almost only \DATA{} tokens.
At other times, it learned to predict low logit values for
\QUERY{} in $v_{dw}$, which has a similar result. In contrast to removing all three
matrices ($M_d$, $M_k$, and $M_p$), in this case the encoder outputs are weighted by the
softmax values as in Eqs.~\ref{eq:Md},~\ref{eq:Mk}, and~\ref{eq:Mp} before calculating the matrix multiplication (Eq.~\ref{eq:indKP}).
This means that those weights are an essential part of the training.

Ablating the attention had no significant impact on performance, but it slowed
down the speed at which the model was able to learn. Similarly, removing the
$M_{pl}$ multiplication did not change the performance of the model. However, to
fix the data flow, we had to replace it by a linear layer mapping the hidden
representation to the length of the input. Thus the input size of the model
became fixed, so neither longer, nor shorter inputs could be handled.

We introduced and motivated a custom logarithmic softmax function in Section~\ref{sec:indirectModel}.
To determine its importance, we checked the effect of replacing it with the standard softmax function.
In this case, the performance dropped significantly for longer examples, and the learning became slower.
This result confirms our motivation: a (custom) logarithmic softmax provides better gradient flow in the middle of an architecture compared to a vanilla softmax function.

\begin{table}[h!]
	\centering
	\begin{tabular}{l c c c c c c}
		w/o & \multicolumn{2}{c}{\dsITen{}} & \multicolumn{2}{c}{\dsITwenty{}} &  \multicolumn{2}{c}{\dsIForty{}}\\
		& TA & WEA & TA & WEA & TA & WEA\\
		\midrule
		base & 100\% & 100\% & 99.99\% & 99.99\% & 100\% & 100\% \\
		\midrule
		fixed-vocab. & 48.94\% & 0\% & - & - & - & - \\
		$M_d$, $M_k$, $M_p$ & 33.34\% & 0\% & - & - & - & - \\
		$M_d$ & 99.99\% & 99.95\% & - & - & - & - \\
		attn. & 100\% & 100\% & 100\% & 100\% & 99.99\% & 99.96\% \\
		$M_{pl}$ & 100\% & 100\% & - & - & - & - \\
		MLogSoft. & 97.26\% & 74.62\% & 13.40\% & 0\% & - & - \\
	\end{tabular}
	\caption{\label{tab:indirectWO} The results of \permneti{} after removing components from it.}
\end{table}

\section{Conclusion}
\label{sec:conclusion}

In this work, we focused on the resolution and rewriting of references in
program code. To achieve this, we abstracted reference rewriting and introduced
the direct and indirect indexing by permutation problems. We created different
architectures (described in Section~\ref{sec:directModel} and \ref{sec:indirectModel}) for handling these problems
(named \permnetd{} and \permneti{}). We compared them to well-known
architectures on synthetic benchmarks. These measurements show that both
\permnetd{} and \permneti{} can outperform previous work both in performance and
stability.

To show the scalability of the presented architectures, we measured their
performance on examples ten times longer than the baselines could handle. We
also tested if \permneti{} can handle arbitrary indexing. For this, we created
the \DICT{} benchmark, containing examples where the \KEY{} tokens can be found
multiple times in the \QUERY{}, on which \permneti{} performs similarly well.

We tested our architecture created for indirect indexing by permutation on a
real\nobreakdash-world problem. We used \permneti{} as an extension to a model for a
decompilation task~\cite{copyDecomp}. The proportion of incorrectly decompiled
code decreased by 42\% compared to the original architecture.

In the future, we are interested in evaluating our architectures on other
real\nobreakdash-world tasks and exploring further generalizations such as allowing multiple
\DATA{} tokens per \KEY{}. Another direction would be to have a different number
of \DATA{} and \INDEX{}\nobreakdash/\QUERY{} tokens. This would be a further generalization
of \DICT{} (where we allowed duplicate \QUERY{} tokens). Another interesting problem
to address is resolving deeper reference chains, e.g. resolving a reference
given by another reference resolution. This would be needed for some
real-world problems like constant propagation.

\section*{Acknowledgements}\label{sec:ackn}

Supported by the EKÖP-25 and EKÖP-KDP-24 University Excellence Scholarship Programs of the Ministry for Culture
and Innovation from the source of the National Research, Development and Innovation Fund.

\newpage

\bibliography{main}

\newpage
\appendix

\section{A real-world example}\label{sec:realWorldExample}

Below is an assembly code snippet corresponding to a \textit{C} switch statement.

\begin{lstlisting}[language=c++]
	// in case of 87 as an argument, jump to .L2 label
	// here ".L2" is the key and "87" is the data
	cmp QWORD PTR -8 [ rbp ] 87
	je .L2
	cmp QWORD PTR -8 [ rbp ] 87
	jg .L3
	cmp QWORD PTR -8 [ rbp ] -30
	je .L4
	cmp QWORD PTR -8 [ rbp ] -30
	jg .L3
	cmp QWORD PTR -8 [ rbp ] -46
	je .L5
	cmp QWORD PTR -8 [ rbp ] -46
	jg .L3
	cmp QWORD PTR -8 [ rbp ] -62
	je .L6
	cmp QWORD PTR -8 [ rbp ] -62
	jg .L3
	cmp QWORD PTR -8 [ rbp ] -72
	je .L7
	cmp QWORD PTR -8 [ rbp ] -66
	je .L8
	jmp .L3
	.L6 :
	<inner-block-1>
	.L5 :
	<inner-block-2>
	.L4 :
	<inner-block-3>
	// at the jump label .L2 run the following inner-block
	// here we want to refer to the data corresponding to ".L2"
	.L2 :
	<inner-block-4>
	.L7 :
	<inner-block-5>
	.L8 :
	<inner-block-6>
	.L3 :
\end{lstlisting}

\begin{lstlisting}[language=c]
	switch (y) {
		// the order of the inner blocks is the same as in the assembly
		// however, the case labels are rearranged in the assembly code
		case -62: <inner-block-1>
		case -46: <inner-block-2>
		case -30: <inner-block-3>
		case  87: <inner-block-4>
		case -72: <inner-block-5>
		case -66: <inner-block-6>
	}
\end{lstlisting}

\newpage
\section{Hyperparameters}\label{hyperparams}

\begin{table}[!h]
	\centering
	\begin{tabular}{c | c c | c c}
		\toprule
		\multirow{3}{*}{Name} & \multicolumn{4}{c}{Dataset} \\
		& \multicolumn{2}{c}{Direct} & \multicolumn{2}{c}{Indirect} \\
		& \dsDOneToTen{} & \dsDTen{} & \dsIOneToTen{} & \dsITen{} \\
		\midrule
		Training data count & \num{103515} & \num{90000} & \num{90292} & \num{90000} \\
		Test data count & \num{11502} & \num{10000} & \num{10033} & \num{10000} \\
		Embedding size & 128 & 128 & 128 & 128 \\
		Hidden size & 1024 & 1024 & 1024 & 1024 \\
		Learning rate & 0.0003 & 0.0003 & 0.0003 & 0.0003 \\
		Batch size & 4096 & 4096 & 4096 & 4096 \\
		Epoch count & 200 & 200 & 200 & 200 \\
		\bottomrule
	\end{tabular}
	\caption{\label{tab:hyper_gru}Hyperparameters of \GRU{} \cite{gru}}
\end{table}

\begin{table}[!h]
	\centering
	\begin{tabular}{c | c c c | c c}
		\toprule
		\multirow{3}{*}{Name} & \multicolumn{5}{c}{Dataset} \\
		& \multicolumn{3}{c}{Direct} & \multicolumn{2}{c}{Indirect} \\
		& \dsDOneToTen{} & \dsDTen{} & \dsDTwenty{} & \dsIOneToTen{} & \dsITen{} \\
		\midrule
		Training data count & \num{102960} & \num{90000} & \num{90000} & \num{39329} & \num{180000} \\
		Test data count & \num{11440} & \num{10000} & \num{10000} & \num{4370} & \num{20000} \\
		Embedding size & 256 & 128 & 256 & 256 & 256 \\
		Hidden size & 1024 & 1024 & 1024 & 1024 & 1024 \\
		Learning rate & 0.0003 & 0.0003 & 0.0003 & 0.0003 & 0.0003 \\
		Batch size & 1024 & 512 & 512 & 1024 & 1024 \\
		Epoch count & 200 & 200 & 200 & 800 & 200 \\
		\bottomrule
	\end{tabular}
	\caption{\label{tab:hyper_attn}Hyperparameters of attention mechanism~\cite{bahdanau}}
\end{table}

\begin{table}[!h]
	\centering
	\begin{tabular}{c | c c c | c c c}
		\toprule
		\multirow{3}{*}{Name} & \multicolumn{6}{c}{Dataset} \\
		& \multicolumn{3}{c}{Direct} & \multicolumn{3}{c}{Indirect} \\
		& \dsDOneToTen{} & \dsDTen{} & \dsDTwenty{} & \dsIOneToTen{} & \dsITen{} & \dsITwenty{} \\
		\midrule
		Training data count & \num{100000} & \num{100000} & \num{100000} & \num{100000} & \num{100000} & \num{100000} \\
		Test data count & \num{200017} & \num{200000} & \num{200000} & \num{209600} & \num{200000} & \num{200000} \\
		Embedding size & 64 & 64 & 64 & 64 & 128 & 128 \\
		Hidden size & 512 & 512 & 512 & 512 & 512 & 512 \\
		Learning rate & 0.0001 & 0.0001 & 0.0001 & 0.0001 & 0.0001 & 0.0001 \\
		Batch size & 512 & 512 & 512 & 512 & \num{2048} & \num{2048} \\
		Epoch count & 200 & 200 & 200 & 200 & 100 & 200 \\
		\bottomrule
	\end{tabular}
	\caption{\label{tab:hyper_copy}Hyperparameters of copying mechanism~\cite{copy} (CopyNet)}
\end{table}

\begin{table}[!h]
	\centering
	\begin{tabular}{c | c c | c c}
		\toprule
		\multirow{3}{*}{Name} & \multicolumn{4}{c}{Dataset} \\
		& \multicolumn{2}{c}{Direct} & \multicolumn{2}{c}{Indirect} \\
		& \dsDOneToTen{} & \dsDTen{} & \dsIOneToTen{} & \dsITen{} \\
		\midrule
		Training data count & \num{152640} & \num{90000} & \num{90292} & \num{90000} \\
		Test data count & \num{16960} & \num{10000} & \num{10033} & \num{10000} \\
		Embedding size & 256 & 256 & 256 & 256 \\
		Hidden size & 1024 & 1024 & 1024 & 1024 \\
		Number of layers & 4 & 4 & 4 & 4 \\
		Number of heads & 12 & 12 & 12 & 12 \\
		Learning rate & 0.001 & 0.001 & 0.001 & 0.001\\
		Batch size & 256 & 256 & 256 & 256 \\
		Epoch count & 500 & 500 & 500 & 500 \\
		\bottomrule
	\end{tabular}
	\caption{\label{tab:hyper_transformer}Hyperparameters of transformer~\cite{attention}}
\end{table}

\begin{table}[!h]
	\centering
	\begin{tabular}{c | c c c | c c c}
		\toprule
		\multirow{3}{*}{Name} & \multicolumn{6}{c}{Dataset} \\
		& \multicolumn{3}{c}{Direct} & \multicolumn{3}{c}{Indirect} \\
		& \dsDOneToTen{} & \dsDTen{} & \dsDTwenty{} & \dsIOneToTen{} & \dsITen{} & \dsITwenty{} \\
		\midrule
		Training data count & \num{100000} & \num{100000} & \num{100000} & \num{100000}& \num{100000} & \num{100000} \\
		Test data count & \num{100000} & \num{100000} & \num{100000} & \num{100000} & \num{100000} & \num{100000} \\
		Embedding size & 64 & 64 & 64 & 128 & 128 & 128 \\
		Hidden size & 32 & 32 & 32 & 256 & 32 & 128 \\
		Learning rate & 0.001 & 0.001 & 0.001 & 0.0003 & 0.003 & 0.0003 \\
		Batch size & \num{16384} & \num{16384} & \num{16384} & \num{4096} & \num{8192} & \num{4096} \\
		Epoch count & \num{1000} & \num{1000} & \num{1000} & 100 & 100 & 100 \\
		\bottomrule
	\end{tabular}
	\caption{
		\label{tab:hyper_perm}Hyperparameters of \permnetd{} (described in Section~\ref{sec:directModel}) and \permneti{} (described in Section~\ref{sec:indirectModel}).
	}
\end{table}

\begin{table}[!h]
	\centering
	\begin{tabular}{c | c c | c c c }
		\toprule
		\multirow{3}{*}{Name} & \multicolumn{5}{c}{Dataset} \\
		& \multicolumn{2}{c}{Direct} & \multicolumn{3}{c}{Indirect} \\
		& \dsDForty{} & \dsDHundred{} & \dsIForty{} & \dsIHundred{} & \DICT{} \\
		\midrule
		Training data count & \num{100000} & \num{100000} & \num{100000} & \num{100000} & \num{100000} \\
		Test data count & \num{100000} & \num{100000} & \num{100000} & \num{100000} & \num{100000} \\
		Embedding size & 128 & 256 & 128 & 128 & 128 \\
		Hidden size & 64 & 128 & 256 & 512 & 64 \\
		Learning rate & 0.0003 & 0.0001 & 0.0003 & 0.00005 & 0.001 \\
		Batch size & \num{4096} & 2048 & \num{1024} & 512 & \num{16384} \\
		Epoch count & \num{1000} & \num{2000} & 100 & 100 & 100 \\
		\bottomrule
	\end{tabular}
	\caption{
		\label{tab:hyper_long}Hyperparameters of \permnetd{} and \permneti{} tested for scalability and \DICT{} benchmarks (described in Section~\ref{sec:datasets}).
	}
\end{table}

\end{document}